\documentclass[10pt,twocolumn,letterpaper]{article}

\usepackage{iccv}
\usepackage{times}
\usepackage{epsfig}
\usepackage{graphicx}
\usepackage{amsmath}
\usepackage{amssymb}
\usepackage{mathtools}
\usepackage[pagebackref=true,breaklinks=true,letterpaper=true,colorlinks,bookmarks=false]{hyperref}

\iccvfinalcopy 

\def\iccvPaperID{4103}

\ificcvfinal\fi

\begin{document}

\title{Graph-Based 3D Multi-Person Pose Estimation Using Multi-View Images}

\author{Size Wu$^{1,3}$ \quad Sheng Jin$^{2,3}$ \quad Wentao Liu$^{3}$\thanks{Corresponding author.} \quad Lei Bai$^{4}$ \quad Chen Qian$^{3}$ \quad Dong Liu$^{1}$ \quad Wanli Ouyang$^{4}$  \\
$^{1}$ University of Science and Technology of China \quad
$^{2}$ The University of Hong Kong \\ $^{3}$ SenseTime Research and Tetras.AI \quad $^{4}$ The University of Sydney \\
\tt\small wsz327471010@mail.ustc.edu.cn  \quad \{jinsheng, liuwentao, qianchen\}@sensetime.com \\ 
\tt\small baisanshi@gmail.com \quad dongeliu@ustc.edu.cn   \quad wanli.ouyang@sydney.edu.au 
}

\maketitle
\ificcvfinal\thispagestyle{empty}\fi

\begin{abstract}
   This paper studies the task of estimating the 3D human poses of multiple persons from multiple calibrated camera views. Following the top-down paradigm, we decompose the task into two stages, \ie person localization and pose estimation. Both stages are processed in coarse-to-fine manners. And we propose three task-specific graph neural networks for effective message passing. For 3D person localization, we first use Multi-view Matching Graph Module (MMG) to learn the cross-view association and recover coarse human proposals. The Center Refinement Graph Module (CRG) further refines the results via flexible point-based prediction. For 3D pose estimation, the Pose Regression Graph Module (PRG) learns both the multi-view geometry and structural relations between human joints. Our approach achieves state-of-the-art performance on CMU Panoptic and Shelf datasets with significantly lower computation complexity.
\end{abstract}

\section{Introduction}
The task of estimating 3D human poses of multiple persons from multiple views is a long-standing problem. It has attracted increasing attention for its wide range of applications, \eg sports broadcasting~\cite{bridgeman2019multi} and retail analysis~\cite{tu2020voxelpose}. 

\begin{figure}[t]
	\centering
	\includegraphics[width=0.45\textwidth]{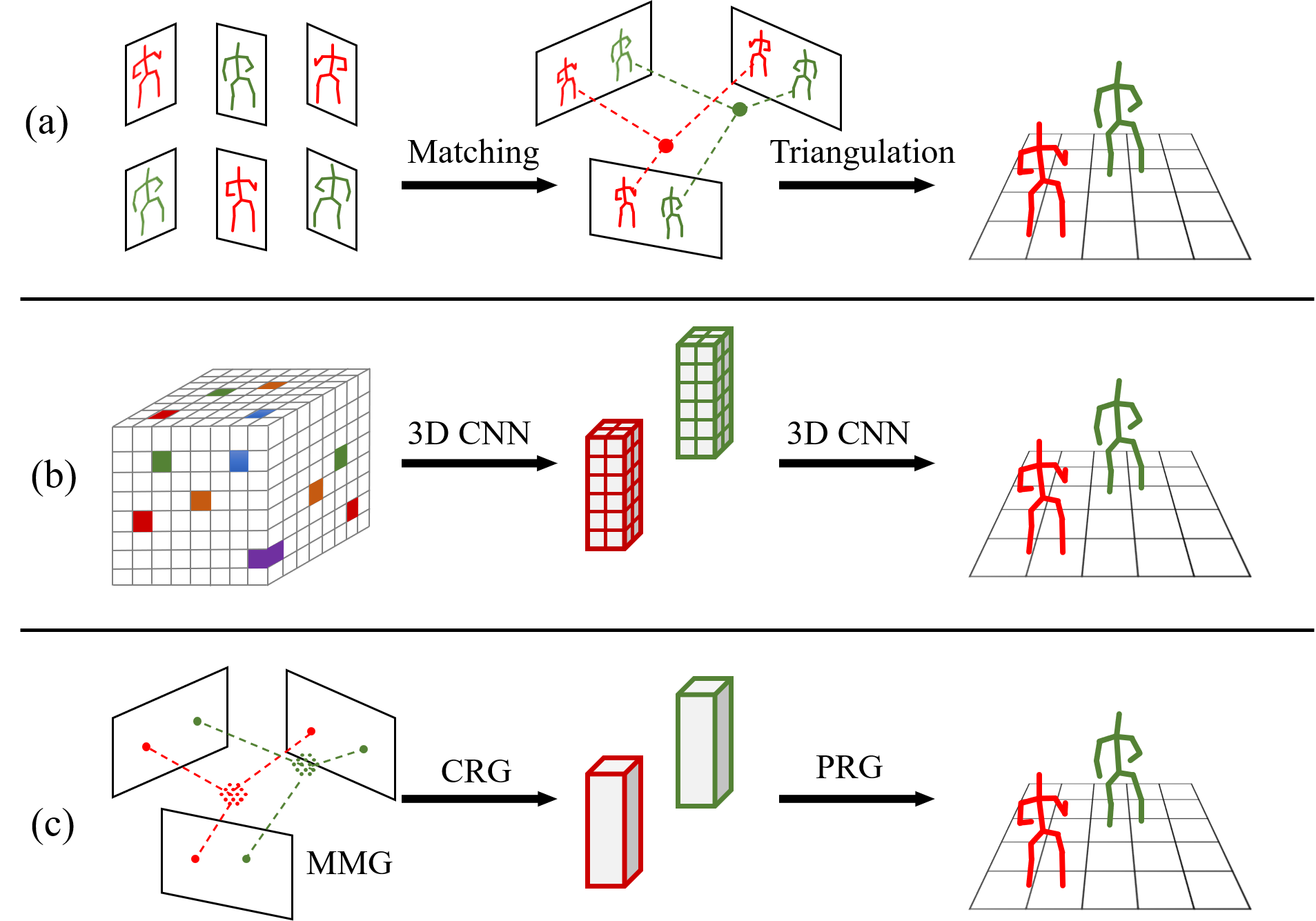}
	\caption{Overview of mainstream multi-view 3D pose estimation frameworks. (a) 2D-to-3D lifting-based approaches (b) Direct 3D pose estimation approaches. (c) Our approach applies graph-based matching algorithm to detect human centers, and applies a graph-based pose refinement model to effectively utilize both geometric cues and human structural prior to achieve better performance. }
	\label{fig:framework_compare}
\end{figure}

Recent research on 3D multi-person pose estimation using multi-view images generally follows two streams: 2D-to-3D lifting-based approaches and direct 3D estimation approaches. As shown in Figrue~\ref{fig:framework_compare}(a), 2D-to-3D lifting approaches~\cite{belagiannis20143d, belagiannis20153d} first estimate 2D joints in each view through monocular pose estimator, then associate 2D poses across views, and finally lift the matched 2D single-view poses to 3D via triangulation~\cite{andrew2001multiple} or Pictorial
Structure Models (PSM)~\cite{dong2019fast}. Such approaches are generally efficient and are the de-facto standard when seeking real-time performance~\cite{remelli2020lightweight}. However, the 3D reconstruction accuracy is limited by the 2D pose estimation, which is not robust to occlusion. As shown in Figure~\ref{fig:framework_compare}(b), direct 3D approaches~\cite{tu2020voxelpose} construct the discretized 3D volumetric representations~\cite{pavlakos2018ordinal,pavlakos2017coarse} by gathering multi-view features and directly operate in the 3D space. 
Such approaches avoid making incorrect decisions in 2D camera views. However, their computation cost increases cubically with the size of the space. They also suffer the quantization errors caused by space discretization~\cite{tu2020voxelpose}.

As shown in Figure~\ref{fig:framework_compare}(c), we combine the virtues of both approaches by adopting 2D-to-3D lifting for efficient \emph{3D human center detection} in the first stage, and direct 3D estimation approach for accurate \emph{single-person 3D pose estimation} in the second stage. To strike a balance between accuracy and efficiency, both stages are processed in coarse-to-fine manners with task-specific graph neural networks.

For coarse-level 3D human center detection in the first stage, we 
generate coarse human center predictions via multi-view matching. Previous methods perform association across views by multi-view geometric constraints~\cite{kadkhodamohammadi2021generalizable} and appearance similarity~\cite{dong2019fast}. However, their matching criteria are hand-crafted and not learnable, which may suffer from tedious hyper-parameter tuning and inaccurate matching results. To solve this problem, we propose the Multi-view Matching Graph Module (MMG) to \emph{learn from data} to match people across views by considering both the visual and geometric cues. It also captures the relationship among multiple views to make more reliable predictions.

For fine-level 3D human center detection in the first stage, we propose a graph-based point predictor, \ie Center Refinement Graph Module (CRG), to refine the coarse human center locations. Previous works~\cite{amin2013multi,bridgeman2019multi,pavlakos2017coarse,pavlakos2018ordinal,tu2020voxelpose} mostly discretize the space into voxels and operate on a regular grid. CRG instead adopts implicit field representations~\cite{kirillov2020pointrend,saito2019pifu,saito2020pifuhd} and directly operates on the continuous 3D space to predict whether a point is a human center or not. It gives us the flexibility to balance between accuracy and speed, by sampling with arbitrary step sizes. Additionally, we propose to use graph models to learn to fuse multi-view features, which are not well-exploited in literature.

For coarse-level single-person pose estimation, we simply use an off-the-shelf pose estimator to generate initial 3D poses based on the detected human proposals. For fine-level single-person pose estimation, we propose the Pose Regression Graph Module (PRG) to refine the initial 3D poses, by exploiting both the spatial relations between body joints and the geometric relations across multiple views. 

The three graph modules can alleviate the aforementioned weakness caused by inaccurate 2D detection or space discretization and improve the pose estimation accuracy.

Our main contributions can be summarized as follows:
\begin{itemize}
\item To the best of our knowledge, this is the first attempt of using task-specific graph neural networks for multi-view 3D pose estimation. We propose a novel coarse-to-fine framework that significantly outperforms the previous approaches both in accuracy and efficiency.

\item We propose Multi-view Matching Graph Module (MMG) to significantly improve the performance of multi-view human association via learnable matching.

\item We propose Center Refinement Graph Module (CRG) for point-based human center refinement, which effectively aggregates multi-view features via graph neural networks, and adaptively samples points to achieve more efficient and accurate localization.

\item We propose a powerful graph-based model, termed Pose Regression Graph (PRG) for 3D human pose refinement. It accounts for both the human body structural information and the multi-view geometry to generate more accurate 3D human poses.

\end{itemize}

\section{Related Work}

\subsection{Single-view 3D pose estimation}

For \emph{single-person 3D pose estimation} from a monocular camera, we briefly classify the existing works into three categories: (1) from 2D poses to 3D poses~\cite{chen20173d,martinez2017simple, zhang2020learning} (2) jointly learning 2D and 3D poses~\cite{nie2017monocular,pavlakos2018ordinal}, and (3) directly regressing 3D poses~\cite{pavlakos2017coarse, pymaf2021} from images. They have shown remarkable results in reconstructing 3D poses, which motivates more research efforts on the more challenging multi-person tasks. \emph{Multi-person 3D pose estimation} from a single RGB image generally follows two streams: top-down and bottom-up. Top-down approaches~\cite{dabral2018learning,moon2019camera,zanfir2018monocular} first use a human detector to produce human locations and then apply single-person pose estimation for each detected person. Bottom-up approaches~\cite{mehta2018single,zanfir2018deep} directly localize keypoints of all people and perform keypoint-to-person association. 

Single-view 3D pose estimation has achieved significant progress in recent years. However, inferring 3D poses from a single view is an ill-posed problem. And its reconstruction accuracy is not comparable with that of the multi-view approaches.

\subsection{Multi-view 3D pose estimation}

We mainly focus on the multi-person 3D pose estimation from multiple views. Existing approaches can be mainly categorized into 2D-to-3D pose lifting approaches~\cite{amin2013multi,belagiannis20143d,belagiannis20153d,bridgeman2019multi,dong2019fast,ershadi2018multiple,huang2020end,lin2021multi,zhang20204d} and direct 3D pose estimation approaches~\cite{tu2020voxelpose}. 

\textbf{2D-to-3D lifting approaches}~\cite{amin2013multi,belagiannis20143d,belagiannis20153d,bridgeman2019multi,dong2019fast,ershadi2018multiple} first estimate 2D joints of the same person in each view through monocular pose estimator, then lift the matched 2D single-view poses to 3D locations. Belagiannis \etal~\cite{belagiannis20143d,belagiannis20153d} first extends 2D PSM to 3D Pictorial Structure Model (3DPS) to encode body joint locations and pairwise relations in between. Other works~\cite{bridgeman2019multi,huang2020end} first solve multi-person 2d pose detection and associate poses in multiple camera views. The 3D poses are recovered using triangulation~\cite{bridgeman2019multi} or single-person 3D PSM~\cite{dong2019fast}. Concurrently Lin \etal~\cite{lin2021multi} propose to use 1D convolution to jointly
address the cross-view fusion and 3D pose reconstruction based on plane sweep stereo. However, such approaches heavily rely on 2D detection results, and the gross errors in 2D may largely degrade 3D reconstruction. In comparison, our approach makes predictions in a coarse-to-fine manner. It models the interaction between multiple camera views using graph neural networks, which are much more efficient and accurate. 

\textbf{Direct 3D pose estimation approaches}~\cite{tu2020voxelpose} discretize the 3D space with volumetric representation and gather features from all camera views via multi-view geometry. Tu \etal proposes to solve multi-person multi-view 3D pose estimation following the top-down paradigm. Specifically, it first discretizes 3D space with voxels and intensively operates on 3D space via 3DCNN to give human proposals. For each human proposal, another 3DCNN is applied to recover 3D human poses. Such approaches reliably recover 3D poses but are computationally demanding. In comparison, our approach introduces MMG to significantly reduce the searching space using the multi-view geometric cues. Combined with point-based predictor CRG, we achieve higher accuracy with less computation complexity. 

Aggregating features from arbitrary views is important but not well-exploited in literature. Traditional methods aggregate multi-view features by concatenation or average-pooling~\cite{tu2020voxelpose}. Feature concatenation can hardly generalize to different camera settings by design. Average-pooling is permutation invariant but ignores the relations between views. In this paper, we propose a novel graph neural network model to learn to combine geometric knowledge with the corresponding 2D visual features from different views.

\subsection{Graph Neural Networks}

Graph Convolutional Networks (GCN) generalizes convolutional neural networks to handle graphic data. GCNs have shown effectiveness in message passing, and global relations modeling in various tasks, \eg action recognition~\cite{yan2018spatial} and tracking~\cite{gao2019graph}. Recent GCNs can be categorized into spectral approaches~\cite{bruna2013spectral,kipf2016semi} and spatial approaches~\cite{duvenaud2015convolutional,wang2019dynamic}. In this paper, we use spatial approaches for better efficiency and generalizability.

Recently, GCN have shown effectiveness in modeling human body structure for \emph{single-view} 2D human pose estimation. Zhang \etal~\cite{zhang2019human} proposes to use PGNN to learn the structured representation of keypoints for 2D single-person pose estimation. Qiu \etal~\cite{qiu2020peeking} proposes OPEC-Net to handle occlusions for 2D top-down pose estimation. Jin \etal~\cite{jin2020differentiable} proposes the hierarchical graph grouping module to learn to associate joints for 2D bottom-up pose estimation. There are also works for \emph{single-view} single-person 3D pose estimation. Zhao \etal~\cite{zhao2019semantic} proposes SemGCN to capture both local and global semantic relationships between joints. Zou \etal~\cite{zou2020high} proposes to capture the long-range dependencies via high-order graph convolution. 

We propose to use graph-based models to learn to aggregate features from multiple camera views via multi-view geometry, which was not investigated in existing GCN works. In Pose Refinement Graph Module (PRG), both the body structure priori and the geometric correspondence of multiple views are encoded for more robust and accurate human pose estimation. Moreover, we propose EdgeConv-E, a variant of EdgeConv~\cite{wang2019dynamic}, to explicitly incorporate geometric correspondence as the edge attributes in GCN.

\subsection{Implicit Field Representations}

Most 3D multi-view pose estimators~\cite{amin2013multi,bridgeman2019multi,pavlakos2018ordinal,pavlakos2017coarse,tu2020voxelpose} use 3D volumetric representations, where 3D space is discretized into regular grids. However, constructing a 3D volume suffers from the cubic scaling problem. This limits the resolution of the volumetric representations, leading to large quantization errors. Using finer grids can improve the performance, but it incurs prohibitive memory costs and computation complexity.

Recently, implicit neural representation or implicit field~\cite{chen2019learning,mescheder2019occupancy,saito2019pifu,saito2020pifuhd} have become popular. Such approaches learn 3D reconstruction in \emph{continuous} function space. Kirillov \etal proposes PointRend~\cite{kirillov2020pointrend} to select a set of points at which to make predictions for instance segmentation. Inspired by PointRend~\cite{kirillov2020pointrend}, we propose Center Refinement Graph (CRG), a point-based predictor, to operate on continuous 3D space in a coarse-to-fine manner. We are able to achieve higher accuracy with significantly lower computation complexity. 

\section{Method}

\begin{figure*}[t]
	\centering
	\includegraphics[width=0.99\textwidth]{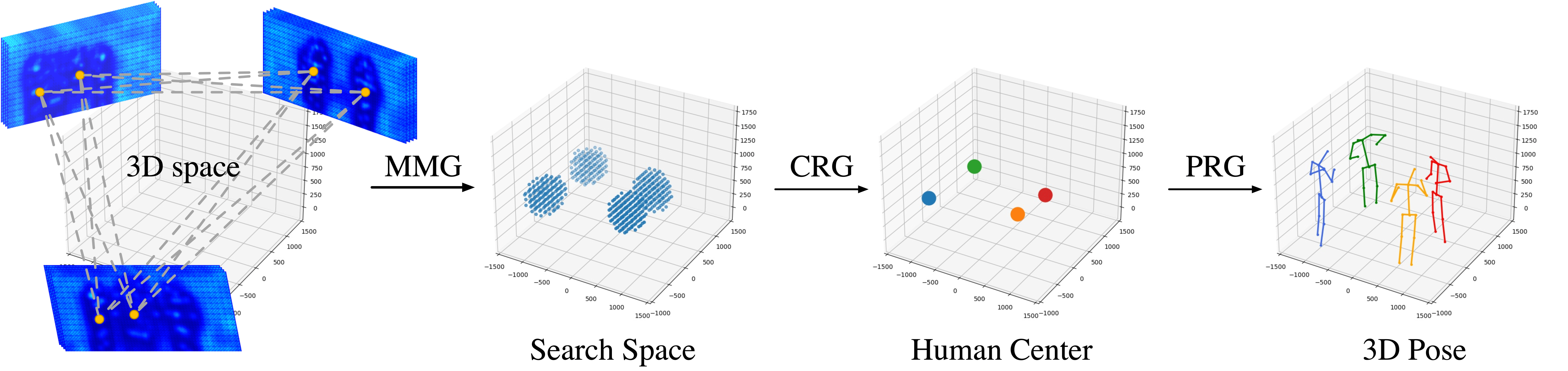}
	\caption{Overview of our approach. The whole pipeline follows the top-down paradigm. It first applies Multi-view Matching Graph Module (MMG) to obtain coarse human center candidates, which are used to limit the search space. Center Refinement Graph Module (CRG) adaptively performs point-based prediction in the search space for more accurate human detection. Finally, Pose Regression Graph Module (PRG) is applied to each detected human proposal to predict the 3D poses in a coarse-to-fine manner.}
	\label{fig:pipeline}
\end{figure*}

\subsection{Overview}

We directly use the same pre-trained 2D bottom-up pose estimator from Tu \etal~\cite{tu2020voxelpose} to localize 2D human centers in each camera view and to provide feature maps for our task-specific GCNs.

To predict the 3D human centers from 2D locations, we propose Multi-view Matching Graph Module (MMG) to match the centers from different camera views corresponding to the same person. Then we obtain coarse 3D human center locations from the matching results via simple triangulation~\cite{andrew2001multiple}. The coarse center candidates are further refined by the Center Refinement Graph Module (CRG).

After 3D human centers are predicted, we follow Tu~\etal~\cite{tu2020voxelpose} to generate 3D bounding boxes with the fixed orientation and size, and apply the 3D pose estimator~\cite{tu2020voxelpose} to generate initial 3D poses. To improve the pose estimation accuracy, the predicted initial 3D poses are further refined by our proposed Pose Regression Graph Module (PRG). 

\subsection{Multi-view Matching Graph Module (MMG)}\label{sec:MMG}

Given the 2D human centers generated by the 2D pose estimator, the proposed Multi-view Matching Graph Module (MMG) aims to match them across different camera views, and lift the 2D human centers to coarse 3D human centers via triangulation~\cite{andrew2001multiple}. We construct a multi-view matching graph, where a vertex represents a human center candidate in a view and an edge represents the connectivity between a pair of human centers in two camera views. The edge connectivity is a binary value in \{0, 1\} representing whether the two corresponding vertices belong to the same person or not. Therefore, the problem of multi-view matching is formulated as the edge connectivity prediction problem. 
Our MMG applies a graph-based model to solve this problem.

The graph model consists of two layers of EdgeConv-E (see Sec.~\ref{sec:edgeconv}) followed by two fully-connected layers. 
 It takes both the vertex features and edge features as input, extracts representative features via message passing, and learns to predict the edge connectivity scores. 

The vertex feature encodes the 2D visual cues which are obtained from the feature maps of the 2D backbone networks. Specifically, the vertex feature vector $\mathbb{R}^{512}$ is extracted at each human center location. The edge feature encodes the pair-wise geometric correspondences of two 2D human centers from two distinct views via epipolar geometry~\cite{andrew2001multiple}. Specifically, we first compute the symmetric epipolar distance~\cite{andrew2001multiple} $d$ between the two centers. Then the correspondence score $s_{corr}$ can be calculated by $s_{corr} = e^{ - m \cdot d }$, where $m$ is a constant and is empirically set to $10.0$ in our implementation. In this way, We explicitly use the geometric correspondence score $s_{corr}$ as the edge feature in MMG.

\subsubsection{Incorporating edge attributes with EdgeConv-E}
\label{sec:edgeconv}
EdgeConv~\cite{wang2019dynamic} is a popular graph convolution prediction to capture local structure and learn the embeddings for the edges. Mathematically, EdgeConv can be represented as:

\begin{equation}\label{eq:edge_conv}
\begin{aligned}
x_v & \doteq \max_{v' \in \mathcal{N}(v)} h_{\mathbf{\theta}} \left( \text{Concat}(x_v, x_{v'} - x_v) \right),
\end{aligned}
\end{equation}
where $x_{v}$ and $x_{v'}$ represent the node features at $v$ and $v'$. `Concat' denotes the feature concatenation operation. $\mathcal{N}(v)$ is the neighbor vertices of $v$. $h_{\mathbf{\theta}}$ is a neural network, \ie a multi-layer perceptron (MLP). 

In standard EdgeConv (Eq.\ref{eq:edge_conv}), the feature aggregation procedure only takes into account the node features $x_{v}$ and the relative relation of two neighboring nodes $(x_{v'} - x_{v})$. It does not explicitly utilize edge attributes for message passing.
Based on EdgeConv~\cite{wang2019dynamic}, we propose EdgeConv-E to explicitly incorporate edge attributes $e_{(v, v')}$ into the aggregation procedure. The propagation rule of EdgeConv-E is illustrated in Eq.\ref{eq:my_edge_conv}.

\begin{equation}\label{eq:my_edge_conv}
\begin{aligned}
x_v & \doteq \max_{v' \in \mathcal{N}(v)} h_{\mathbf{\theta}}\left( \text{Concat}(x_v, x_{v'} - x_v, e_{(v, v'))}\right).
\end{aligned}
\end{equation}

\subsubsection{Training} 

We first construct a multi-view graph, where the vertices are generated using the 2D human centers, and the edges connect each pair of 2D human centers in distinct camera views. The target edge connectivity is assigned $``1"$s for edges connecting the same persons, and $``0"$s otherwise. To avoid overfitting, we augment by adding uniform noises ranging from 0 to 25 pixels to the ground-truth 2D human center coordinates. Binary cross-entropy loss between the predicted and the target edge connectivity is used for training. We adopt Adam optimizer~\cite{kingma2014method} with a learning rate of $10^{-4}$ to train the model for 2 epochs.

\subsection{Center Refinement Graph Module (CRG)}
\label{sec:CRG}

\begin{figure}[t]
	\centering
	\includegraphics[width=0.49\textwidth]{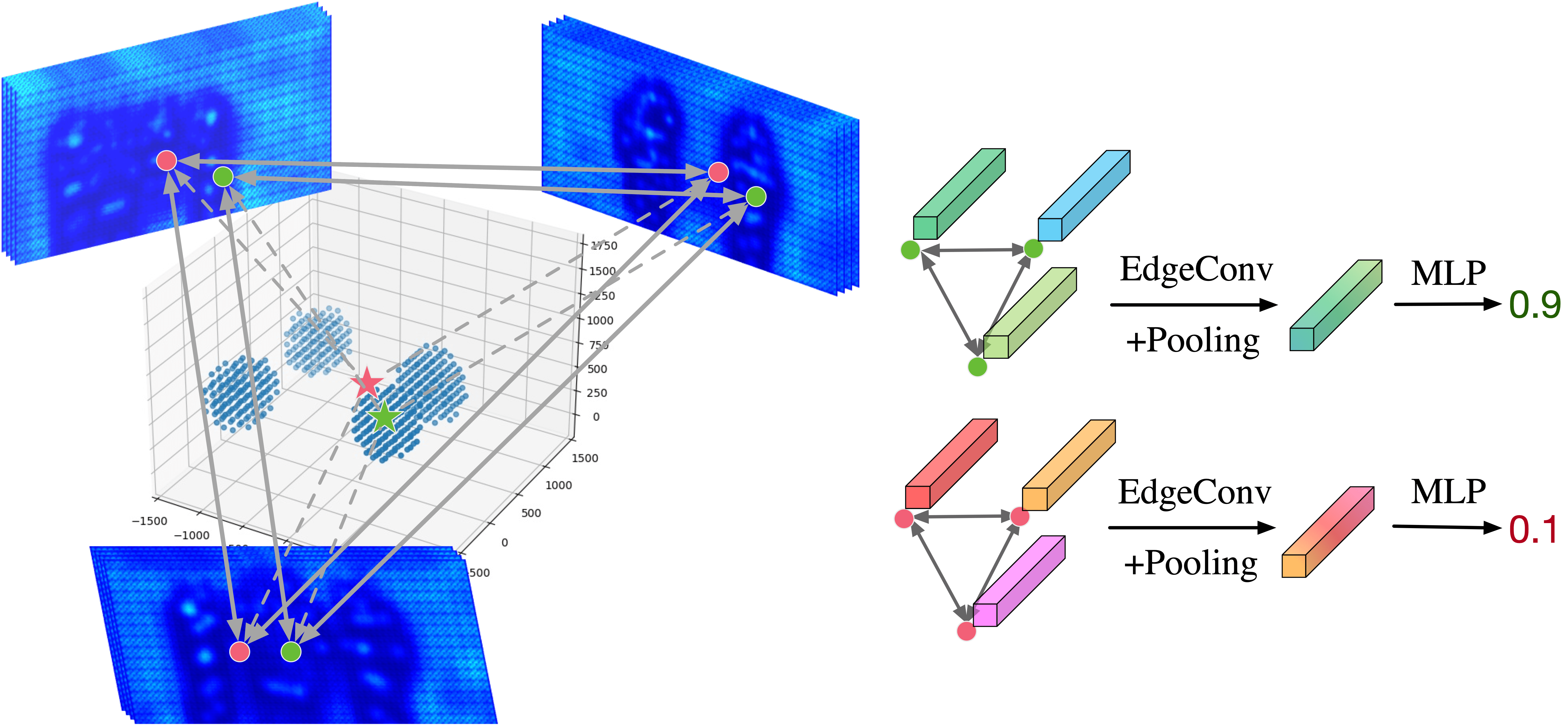}
	\caption{Center Refinement Graph Module (CRG) iteratively applies point-based prediction on selected query points to detect human centers. The graph is constructed by linking the 2D projections of the 3D query in all camera views. Through a few graph convolutions, graph pooling, and MLP, we obtain the confidence score for each proposal.}
	\label{fig:center_graph}
\end{figure}

Center Refinement Graph Module (CRG) is built on top of MMG to refine the 3D human center detection results. CRG adaptively samples query points in the 3D search space, and predicts the possibility of the query point being a human center. It replaces the commonly used volumetric representations with the implicit field representations, which enables querying at any \emph{real-value} point for more flexible search and accurate localization in the 3D space. 

\textbf{Search space.} 
\label{sec:search_space}
Instead of operating on the whole 3D space, we propose to restrict the search space based on the matching results from MMG. For each pair of matched 2D human centers, we recover a coarse 3D human center proposal via triangulation~\cite{andrew2001multiple}. We generate a 3D ball surrounding each 3D human center proposal within a radius of $r_0=\text{300mm}$. The search space (denoted as $\Omega_0$) is thus the union of these 3D balls.

\textbf{Feature extraction.} Each query 3D point is first projected to all 2D camera views to get its corresponding 2D locations. Then the point-wise feature representations of the corresponding 2D point locations are obtained from the 2D feature maps. Features for a real-value 2D location are obtained via bilinear interpolation, using the surrounding four nearest neighbors located on the regular grid. 

We first introduce a baseline model, which concatenates the point-wise features from different views and processes with a learnable multi-layer perceptron (MLP). For each candidate point, the MLP outputs a confidence score of being a human center. We refer to this approach as \emph{MLP-Baseline}. Although intuitive, we argue that this approach is limited for two reasons: (1) it assigns the same weights to all views, and cannot handle occlusion in some viewpoints. (2) it cannot generalize to other camera settings (different number of cameras) by design.

To alleviate these limitations, we propose to use graph neural networks for efficient message passing across views. Our Center Refinement Graph Module (CRG) learns to fuse information from multiple views and verify the proposals from the previous stage. As shown in Figure~\ref{fig:center_graph}, for each 3D query point, we construct a multi-view graph. The vertices represent the 2D projections in each camera view. The vertex features include (1) visual features $\mathbb{R}^{512}$ extracted in the image plane (2) normalized 3D coordinates $\mathbb{R}^{3}$ of the query point. (3) 2D center confidence score from the 2D backbone. The edges densely connect these 2D projections to each other, enabling cross-view feature aggregation. 

Our CRG uses three layers of EdgeConv for cross-view feature message passing, followed by a max-pooling layer for feature fusion and one fully-connected (FC) layer to predict the center confidence score. We use the standard EdgeConv instead of EdgeConv-E, because CRG does not have explicit edge features for aggregation.

\subsubsection{Point Selection}
\label{sec:point_selection}

\textbf{Inference.} Given a search region from MMG, we iteratively search for the human centers in a coarse-to-fine manner. CRG starts with the search space $\Omega_0$ described in Sec.~\ref{sec:search_space}. In the iteration $t$, it uniformly samples query points in the search space, with the step size $\tau_t$. The graph model processes the sampled queries and predicts their possibility of being a human center. The point with the highest confidence score is selected as the refined human center $x_t$. We update the search space for the next iteration, $\Omega_{t+1}$, as the 3D ball subspace surrounding the human center $x_t$ with a radius of $r_{t+1} = r_{t} \cdot \gamma$. We shrink the sampling step size by~\ie $\tau_{t+1} = \gamma' \cdot \tau_{t}$. The iteration continues until the step size reaches the desired precision ($\epsilon$).

\textbf{Complexity analysis.}
In Tu \etal~\cite{tu2020voxelpose}, the search space of human center proposals, as well as the time complexity, is $O(L \times W \times H)$, where $L$, $W$, and $H$ are the size of the 3D space. Applying our proposed MMG and CRG, the size of the search space is significantly reduced to $O(N)$, where $N$ is the number of people. Here we omit the size of the search region of an instance, which is a constant. It is noticeable that the complexity is independent of the size of the space, making it applicable to large space applications, \eg the football field. In the experiments, we set the initial step size $\tau_0 = 200 \text{mm}$, the shrinking factor $\gamma = 0.6$ and $\gamma'=0.25$, the desired precision $\epsilon = 50\text{mm}$. On CMU Panoptic ~\cite{joo2017panoptic} dataset, we record an average of 1,830 queries per frame compared with 128,000 queries taken by Tu~\etal~\cite{tu2020voxelpose}.

\subsubsection{Training} 

The model learns to predict the confidence score for each query point. We develop an effective sampling strategy for selecting training samples to train CRG. Two types of samples are considered for training: positive samples that are located around the ground-truth human centers and negative ones that are far away from human locations. We take positive samples around ground-truth human centers following the Gaussian distributions with the standard deviation $\sigma_{pos} = 400$mm. For negative ones, we take samples uniformly in the entire 3D space. Empirically, the ratio of the number of positive and negative samples is $4:1$.

For a sample located at $\mathbf{X}$, the target confidence score is calculated by
\begin{equation}
    s^*_{conf} = \max_{j=1:N} exp \left\{-\frac{\lVert \mathbf{X} - \mathbf{X}_j^* \rVert_{2}^{2}}{2\sigma^2}\right\},
\end{equation}
where N is the number of human instances and $\mathbf{X}_j^*$ is the 3D coordinate of the center point of person $j$. And $\sigma$ is the standard deviation of the Gaussian distribution, which is set as $\sigma = 200$mm. The training loss of CRG is the $\ell_2$ loss between the predicted and the target confidence score. We adopt Adam optimizer~\cite{kingma2014method} with a learning rate of $10^{-4}$. It takes 4 epochs to reach the best performance.

\subsection{Pose Regression Graph Module (PRG)}

Existing 3D pose regression models produce reliable results on joints that are visible in most views, but will generate inaccurate localization results for occluded joints. Human beings can easily recognize occluded poses, mainly because of their prior knowledge of bio-mechanical body structure constraints and multi-view geometry. The knowledge helps remove ambiguity in localization caused by self-occlusion or inter-occlusion. In light of this, we design the Pose Regression Graph Module (PRG) to learn to refine joint locations considering both the multi-view geometry and structural relations between human joints. 

\begin{figure}[t]
	\centering
	\includegraphics[width=0.49\textwidth]{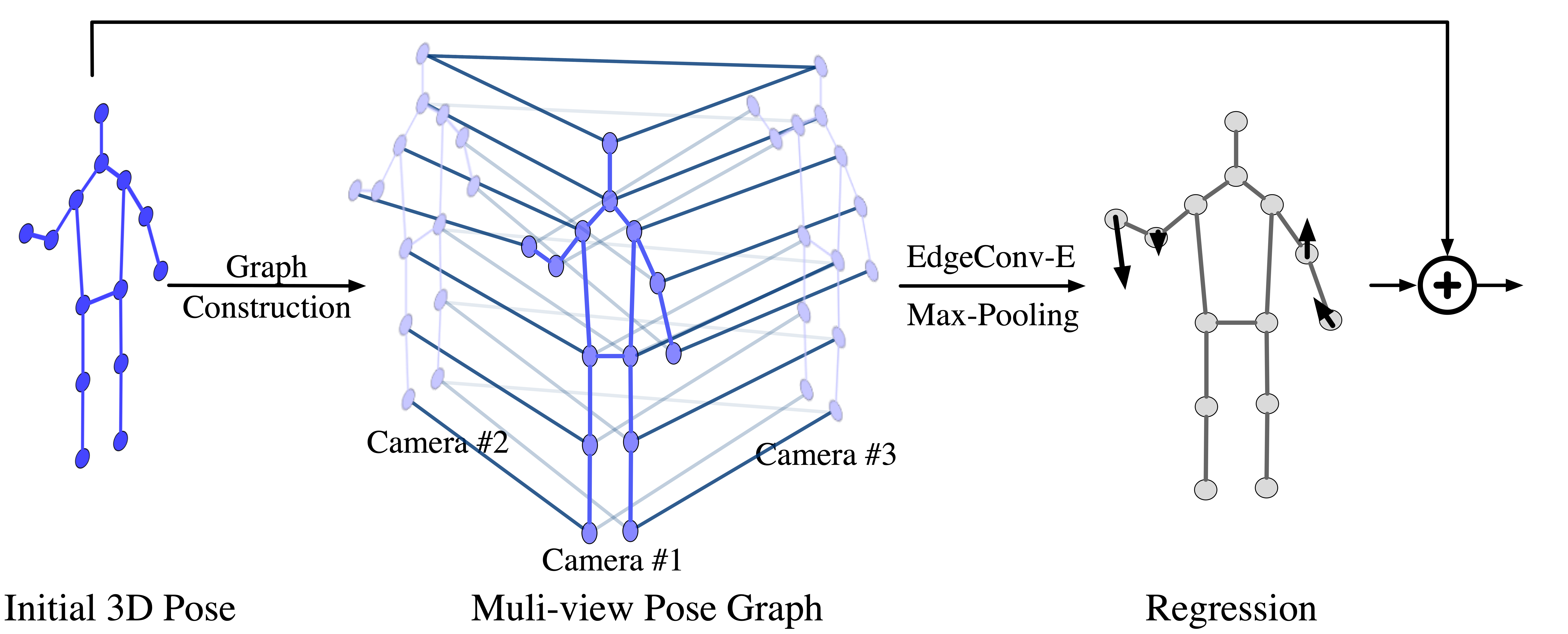}
	\caption{Overview of the 3D pose estimation stage. The initial 3D pose is projected to all camera views to construct the multi-view pose graph. With effective message passing and feature fusion, PRG predicts the regression offsets for 3D pose refinement.}
	\label{fig:pose_graph}
\end{figure}

An overview of the 3D pose estimation stage is illustrated in Figure~\ref{fig:pose_graph}. We applied PRG to each individual to further improve the accuracy. The PRG module takes an initial 3D pose as the input. In our implementation, we simply use the pose regressor of~\cite{tu2020voxelpose} to generate the initial 3D pose. The initial 3D pose is projected to all camera views to obtain multiple 2D poses. 
We construct a multi-view pose graph based on the projected 2D poses in different camera views. The graph predicts the offsets of each keypoint in 3D space, which are added to the initial 3D pose for refinement. 

For the multi-view pose graph, the vertices represent the 2D keypoints in a certain camera view. We concatenate the following features the initialize all the nodes in the graph: 
(1) visual features $\mathbb{R}^{512}$ obtained from the feature maps of the 2D backbone networks at the projected 2D location.
(2) one-hot representation of the joint type $\mathbb{R}^{K}$ (3) normalized initial 3D coordinates $\mathbb{R}^{3}$. 

The multi-view pose graph consists of two types of edges: (1) single-view edges that connect two keypoints of different types in the canonical skeleton structure in a certain camera view. (2) cross-view edges that connect two keypoints of the same type in different views. We use the one-hot feature vector $\mathbb{R}^{2}$ to distinguish these two types of edges. The one-hot edge features are passed to the EdgeConv-E defined by Eq.~\ref{eq:my_edge_conv}.

Our graph model of PRG first uses two consecutive EdgeConv-E layers for message passing between neighboring body joints and multiple camera views. Then a max-pooling layer is applied to aggregate the cross-view features and coarsen the graph. The max pooled features are updated by the following three EdgeConv-E layers via effective information flow between the body joints. Finally, the extracted features are passed to one MLP with two fully-connected (FC) layers to regress a refinement vector for each joint.

\textbf{Training.} The target offset is the difference between the ground-truth 3D pose and the initial 3D pose. We use $\ell_1$ regression loss between the predicted offset and the target offset to train PRG. Note that the loss gradients of PRG can be back-propagated to the 2D backbone network, which will further improve its feature representation ability. We train PRG using the Adam optimizer~\cite{kingma2014method} with a learning rate of $5\times 10^{-5}$. We train it for 4 epochs to obtain the best model.

\section{Experiments}

\subsection{Datasets}

\textbf{CMU Panoptic}~\cite{joo2017panoptic}:
The CMU Panoptic dataset is currently the largest real-world dataset for multi-person 3D pose estimation. It is captured in a studio laboratory, with multiple people doing social activities. In total, it contains 65 sequences (5.5 hours) and 1.5 million of 3D skeletons with 30+ HD camera views. We follow~\cite{tu2020voxelpose,xiang2019monocular} to split the dataset into training and testing subsets. However, we lack the `160906band3' training subset due to broken images. Mean Average Precision (mAP) and mean Average Recall (mAR) are popular metrics for comprehensive evaluation. We calculate mAP and mAR by taking the mean of AP and AR over all the Mean Per Joint Position Error (MPJPE) thresholds (from 25mm to 150mm with a step size of 25mm). We report mAP and mAR along with MPJPE for evaluating the performance of both 3D human center detection and 3D human pose estimation.

\textbf{Shelf}~\cite{belagiannis20143d}:
The Shelf dataset consists of four people disassembling a shelf captured by five cameras. It is challenging due to the complex environment and heavy occlusion. We follow~\cite{belagiannis20143d,dong2019fast,tu2020voxelpose} to prepare the training and testing datasets. Following~\cite{tu2020voxelpose}, we use the same 2D pose estimator trained on the COCO dataset. We follow~\cite{belagiannis20143d,belagiannis20153d,belagiannis2014multiple,dong2019fast,ershadi2018multiple} to use the percentage of correctly estimated parts (PCP3D) to evaluate the estimated 3D poses.

\subsection{Comparisons to the state-of-the-arts}

\begin{table}[h]
    \centering
    \caption{Comparisons to the state-of-the-art approaches on CMU Panoptic dataset~\cite{joo2017panoptic}. The symbol $\uparrow$ means that the higher score the better, while $\downarrow$ means that the lower the better. `*' indicates the mean value of four AP$_{K}$ metrics reported in~\cite{tu2020voxelpose, lin2021multi}. `$\ddagger$' indicates that better 2D pose estimator~\cite{sun2019deep} is used.}
    \scalebox{0.93}{
    \begin{tabular}{c|l|c|c}
\hline
& $\text{mAP}$ $\uparrow$ & $\text{mAR}$ $\uparrow$ & MPJPE $\downarrow$ \\
\hline
Tu \etal~\cite{tu2020voxelpose} & $95.40^{*}$  & -   & 17.68mm \\ 
Tu \etal~\cite{tu2020voxelpose} (reproduce) & 96.73 &  97.56   & 17.56mm \\ 
$^\ddagger$ Lin \etal~\cite{lin2021multi} &$97.68^{*}$& - & 16.75mm \\
Ours&  \textbf{98.10} &   \textbf{98.70}   & \textbf{15.84mm} \\ 
\hline
    \end{tabular}}
    \label{tab:compare_panoptic}
\end{table}

In this section, we compare with the state-of-the-art approaches on CMU Panoptic~\cite{joo2017panoptic} and Shelf~\cite{belagiannis20143d} datasets. 

On CMU Panoptic dataset, we follow~\cite{tu2020voxelpose} to experiment with the five camera setups. To make fair comparisons, we use the same HD camera views (id: 3, 6, 12, 13, 23). As the AP$_{75}$, AP$_{125}$ and mAR are not reported in the original paper of Tu \etal~\cite{tu2020voxelpose}, we reproduce the results by running the publicly available official codes\footnote{https://github.com/microsoft/voxelpose-pytorch} with the recommended hyper-parameters. We find that our re-implementation achieves a slightly better result (17.56mm vs 17.68mm). We show that our approach significantly improves upon Tu \etal~\cite{tu2020voxelpose} on mAP, mAR, and MPJPE. Compared with Tu \etal~\cite{tu2020voxelpose}, our approach has higher accuracy (98.10 mAP vs 96.73 mAP) and also higher recall (98.70 mAR vs 97.56 mAR).
Especially, the MPJPE remarkably decreases from 17.56mm to 15.84mm, demonstrating the effectiveness of our proposed method in reducing the quantization error caused by space discretization. 

The quantitative evaluation results on Shelf~\cite{belagiannis20143d} dataset are presented in Table~\ref{tab:compare_campus}. 
In the experiments, we follow the evaluation protocol of Tu \etal~\cite{tu2020voxelpose}. We show that our approach achieves the state-of-the-art performance.

\begin{table}[h]
    \centering
    \caption{Quantitative comparisons to the state-of-the-art approaches on Shelf~\cite{belagiannis20143d} datasets. The metric is the percentage of correctly estimated parts (PCP3D). `†' means method with temporal information.}
    \scalebox{0.90}{
    \begin{tabular}{c|c|c|c|c}
\hline
Shelf & Actor1 & Actor2 & Actor3 & Average \\ \hline
Belagiannis \etal~\cite{belagiannis20143d}& 66.1 & 65.0 & 83.2 & 71.4\\
†Belagiannis \etal~\cite{belagiannis2014multiple}& 75.0 & 67.0 & 86.0& 76.0\\
Belagiannis \etal~\cite{belagiannis20153d}& 75.3 & 69.7 & 87.6 & 77.5\\
Ershadi \etal~\cite{ershadi2018multiple}& 93.3 & 75.9 & 94.8 & 88.0\\
Dong \etal~\cite{dong2019fast}& 98.8 & 94.1 & 97.8& 96.9\\
Tu \etal~\cite{tu2020voxelpose}& 99.3 & 94.1 & 97.6& 97.0\\
Huang \etal~\cite{huang2020end}&98.8 &96.2 &97.2 & 97.4 \\ 
†Zhang \etal~\cite{zhang20204d} &99.0 &96.2 &97.6 &97.6\\
Ours& 99.3& 96.5 & 97.3& \textbf{97.7}\\ \hline
    \end{tabular}}
    \label{tab:compare_campus}
\end{table}

\begin{figure*}[ht]
	\centering
	\includegraphics[width=0.99\textwidth]{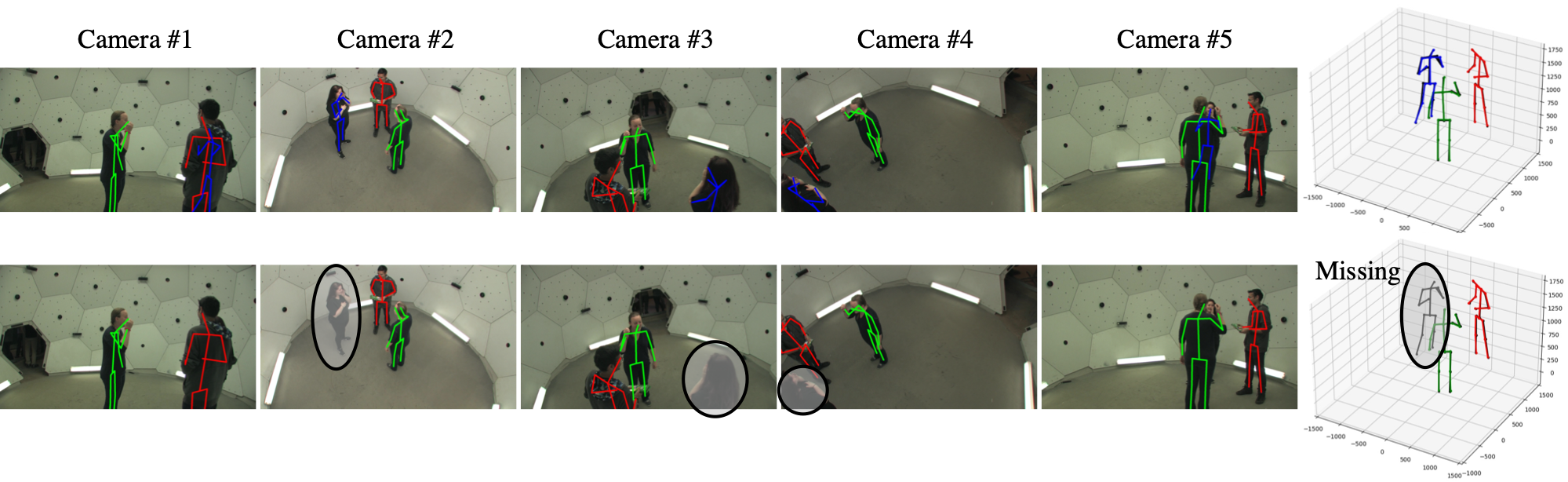}
	\caption{\textbf{Qualitative analysis.} Estimated 3D poses and their 2D projections of ours (the 1st row), and Tu \etal~\cite{tu2020voxelpose} (the 2nd row). 
	The last column illustrates the ground-truth (black) and the predicted 3D poses (red, green, and blue). Missing poses are highlighted with circles.}
	\label{fig:qualitative}
\end{figure*}

\subsection{Ablation study}

In this section, we conduct ablative experiments to analyze each component in our proposed framework in detail. 

\textbf{Effect of MMG.}
In Table~\ref{tab:ablation_MMG_CRG}, we evaluate the performance of Multi-view Matching Graph Module (MMG) on 3D human center detection and 3D human pose estimation. All results use the same 2D detections and 3D human centers are recovered using multi-view triangulation~\cite{andrew2001multiple}. Traditional methods perform association across views using epipolar constraints~\cite{kadkhodamohammadi2021generalizable}. However, they do not generate reliable matching results in occluded scenes. MMG \emph{learns} from data to match people across views.
We observe significant improvement in the matching performance (75.91 mAP vs 61.65 mAP). We also notice that replacing MMG with the ground-truth matching results does not notably improve the human center detection results (78.70 mAP vs 75.91 mAP). This implies that the human association results generated by MMG are already very accurate. 

\textbf{Effect of CRG.}
The Center Refinement Graph Module (CRG) aims at refining the coarse human center predictions. To show the effectiveness of the graph reasoning for human center prediction, we compare CRG with the \emph{MLP-Baseline} introduced in Sec.~\ref{sec:CRG} on CMU Panoptic dataset. For fair comparisons, we make both models share the same input features, and have roughly the same number of parameters. As shown in Table~\ref{tab:ablation_MMG_CRG}, CRG outperforms the \emph{MLP-Baseline} in terms of both human detection accuracy (82.10 mAP vs 81.38 mAP) and 3D human pose estimation accuracy (98.10 mAP vs 97.82 mAP).
This indicates the importance of learning the multi-view relationship via graph-based message passing.

\begin{table}[h]
    \centering
    \caption{Effect of MMG and CRG on human ceter detection and 3D human pose estimation. Pose results in this table are all obtained by PRG. `Epi' means epipolar matching.
    `GT' means using ground-truth matching results. 
    }
    \scalebox{0.85}{
    \begin{tabular}{c|c|c|c|c}
\hline
Method & Center & Pose & Pose & Pose \\
  & mAP $\uparrow$  & mAP $\uparrow$ & mAR $\uparrow$ & MPJPE $\downarrow$ \\
\hline
Epi+Triangulation  & 61.65 & 86.02 & 91.08 & 24.46mm \\
\textbf{MMG}+Triangulation  & 75.91 & 95.11 & 97.60 & 16.99mm \\
GT+Triangulation   & 78.70 & 96.77 & 98.44 & 16.08mm \\
\textbf{MMG}+\emph{MLP-Baseline} & 81.38 & 97.82 & 97.89 & 16.06mm \\
Epi+\textbf{CRG} & 79.80 & 95.68 & 95.68 & 16.03mm \\
\textbf{MMG}+\textbf{CRG} (final)  & \textbf{82.10} & \textbf{98.10} & \textbf{98.70} & \textbf{15.84mm} \\
\hline
    \end{tabular}}
    \label{tab:ablation_MMG_CRG}
\end{table}

\textbf{Effect of PRG.} To analyze the effect of the Pose Regression Graph (PRG), we conduct experiments on CMU Panoptic dataset with multiple initial 3D pose regressors of different accuracy. These models are obtained by varying the granularity of the voxels, \ie $32^3$, $48^3$, and $64^3$. We report the accuracy of the poses before and after the PRG refinement in Table~\ref{tab:init_pose}. Our PRG is a general pose refiner, which can be applied to various pose estimators to consistently improve the 3D pose estimation accuracy. Note that the 3D pose estimator of (c), is from Tu~\etal~\cite{tu2020voxelpose}. 

\begin{table}[h]
    \centering
    \caption{Improvement of 3D pose estimation (MPJPE $\downarrow$) when PRG is applied to different initial 3D pose regressors.}
    \scalebox{1.0}{
    \begin{tabular}{c|c|c|c}
\hline
& Before PRG  & After PRG  & Improvement \\
\hline
(a) & 18.12mm & 16.63mm  & 1.49mm (8.2\%) \\
(b) & 17.78mm & 16.44mm  & 1.34mm (7.5\%) \\
(c) & 17.09mm & 15.84mm  & 1.25mm (7.3\%) \\
\hline
    \end{tabular}}
    \label{tab:init_pose}
\end{table}

\subsection{Qualitative Study}

We qualitatively compare our results with those of Tu \etal~\cite{tu2020voxelpose} in Figure~\ref{fig:qualitative}. In this example, the body of the woman (blue) is only clearly captured by one camera (view \#2), while it is either truncated or occluded in other views. Tu \etal~\cite{tu2020voxelpose} simply averages features from all the views with the same weights. This will make the features unreliable, leading to false negatives (FN). In comparison, our approach learns the multi-view feature fusion via GCN. We obtain more comprehensive features which allows us to make more robust estimation. Our approach also gets fewer false positives (FP) and predicts human poses with higher precision. Please see the supplementary for more examples. 

\subsection{Memory and Runtime Analysis}

\begin{table}[h]
    \centering
    \caption{Memory and runtime analysis on CMU Panoptic dataset. Runtime is tested with one Titan X GPU. $^*$ denotes the cost of processing one person proposal.}
    \scalebox{0.80}{
    \begin{tabular}{c|c|c|c|c|c}
\hline
& CPN~\cite{tu2020voxelpose}  & PRN$^*$~\cite{tu2020voxelpose} & MMG & CRG & PRG$^*$ \\\hline
Memory & 1.10GB & 2.38GB  & 7.10MB & 1.08MB & 20.3MB  \\
Runtime & 26ms & 52ms  &  2.4ms  & 5.6ms & 6.8ms  \\
\hline
    \end{tabular}}
    \label{tab:runtime}
\end{table}

Table~\ref{tab:runtime} reports the memory and runtime on the sequences with 5 camera views on CMU Panoptic dataset. The results are tested on a desktop with one Titan X GPU. Tu~\etal~\cite{tu2020voxelpose} proposes CPN to localize people, and PRN to regress 3D poses. Both of them use volumetric representations, which suffer from large amount of memory. In comparison, the memory cost of our proposed graph neural networks is negligible. Our presented modules are also very efficient. On average, our unoptimized implementation takes only 2.4ms for multi-view matching (MMG) and 5.6ms for finer multi-person human center prediction (CRG). Compared with the CPN in \cite{tu2020voxelpose}, CRG requires tens of fewer sampling queries (1.8K vs 128K) due to smaller searching space. And the time cost of PRG is $6.8$ms for each person.

When using the PRN as the initial pose estimator, our method facilitates the use of fewer bins of the voxel representation. Comparing \#1 and \#4 in Table~\ref{tab:runtime_detail}, our method using $32^3$ bins has about 1/4 computational cost and higher accuracy (1.84mm improvement) than Tu \etal ~\cite{tu2020voxelpose}. Reducing the bins leads to smaller error increase for ours (0.11mm comparing \#2 and \#4), but large error increase for Tu \etal ~\cite{tu2020voxelpose} (1.51mm comparing \#1 and \#3).

\begin{table}[h]
    \centering
    \caption{Runtime comparison. $N$ is the number of persons. `avg is the average runtime (ms) when $N=4$. `\#bins' is the number of bins (voxel granularity) for PRN. }
    \scalebox{0.82}{
    \begin{tabular}{c|c|c|c|c|c}
\hline
\# & Method &\#bins & Computational cost & avg &MPJPE $\downarrow$ \\
\hline
1 & Tu \etal~\cite{tu2020voxelpose} & $64^3$ & $26+52\times N$& 234 & 17.68mm \\
2 & Ours  & $64^3$ & $8 +(52+6.8) \times N$ & 243&15.84mm \\ \hline
3 &Tu \etal~\cite{tu2020voxelpose} & $32^3$ & $26+7.3\times N$& 55& 19.19mm \\
4 &Ours  & $32^3$ & $8 +(7.3+6.8) \times N$ & 64&15.95mm \\
 \hline
    \end{tabular}}
    \label{tab:runtime_detail}
\end{table}

\section{Conclusion}

In this paper, we propose a novel framework for multi-view multi-person 3D pose estimation. We elaborately design three task-specific graph neural network models to exploit multi-view features. We propose Multi-view Matching Graph Module (MMG) and Center Refinement Graph Module (CRG) to detect human centers by proposal-and-refinement, and Pose Regression Graph Module (PRG) to produce accurate pose estimation results. Comprehensive experiments demonstrate that the proposed approach significantly outperforms the previous approaches.

\textbf{Acknowledgement.} We would like to thank Lumin Xu and Wang Zeng for their valuable feedback to the paper. This work is supported by the Australian Research Council Grant DP200103223 and FT210100228, Australian Medical Research Future Fund MRFAI000085, the Natural Science Foundation of China under Grants 62036005 and 62021001, and the Fundamental Research Funds for the Central Universities under contract WK3490000005.

{\small
\bibliographystyle{ieee_fullname}
\bibliography{egbib}
}

\newpage

\appendix
\section*{\Large Appendix}
\setcounter{table}{0}
\renewcommand{\thetable}{A\arabic{table}}
\setcounter{figure}{0}
\renewcommand{\thefigure}{A\arabic{figure}}

\section{Generalization to Different Number of Camera Views}

In this section, we evaluate the generalization to the different number of camera views. Specifically, we train our graph-based models on the five-camera setup (camera id: 3, 6, 12, 13, 23), and directly evaluate these models with different number of camera views, \ie the five-camera setup (camera id: 3, 6, 12, 13, 23), four-camera setup (camera id: 6, 12, 13, 23) and the three-camera setup (camera id: 6, 12, 23). 

Transferring the pre-trained models to a reduced number of camera views is challenging. First, reducing the number of cameras increases the ambiguity of occluded human poses. Second, the information in the fused features is less complete. Third, the feature distribution may vary in different camera setups. We find that Tu \etal~\cite{tu2020voxelpose} does not produce reliable prediction results when transferring to a reduced number of camera views. For example, when the number of camera views ($\#$ Views) is reduced to 3, the mAP drops dramatically from $96.73$ to $68.14$, and the MPJPE increases from $17.56$mm to $37.14$mm. Re-training the models with the test-time camera setups will mitigate this problem (marked with $\dag$). In comparison, our approach can better generalize to different camera setups \emph{without} any fine-tuning. Although reducing the number of camera views will reduce the accuracy, we show that we still achieve reasonably good results, demonstrating that our proposed approach has strong generalization ability. For example, with only 3 camera views, we achieve $91.60$mAP and $94.14$mAR. We also show that our approach consistently outperforms the state-of-the-art approach~\cite{tu2020voxelpose} on generalization to different camera setups. 

\begin{table}[h]
    \centering
    \caption{Generalization to different number of camera views. All results are obtained using ResNet-50 as the backbone. $\uparrow$ means the higher score the better, while $\downarrow$ means the lower the better. $\dag$ means fine-tuning models under the test-time camera setups.}
    \vspace{0.2cm}
    \scalebox{1.0}{
    \begin{tabular}{c|c|c|c|c}
\hline
& \#Views & $\text{mAP}$ $\uparrow$ & $\text{mAR}$ $\uparrow$ & MPJPE $\downarrow$ \\
\hline
Tu \etal~\cite{tu2020voxelpose} & 5 & 96.73 &  97.56   & 17.56mm \\ 
Ours& 5 & \textbf{98.10} &   \textbf{98.70}   & \textbf{15.84mm} \\ \hline
Tu \etal~\cite{tu2020voxelpose} & 4& $94.54$  & 95.97   & 20.06mm \\ 
Tu \etal~\cite{tu2020voxelpose}$^\dag$ & 4 & $95.60$  & 96.80   & 18.63mm \\ 
Ours & 4  & \textbf{97.65} &  \textbf{97.89}   & \textbf{17.87mm} \\ 
\hline
Tu \etal~\cite{tu2020voxelpose} & 3 & $68.14$  & 72.14  & 37.14mm \\ 
Tu \etal~\cite{tu2020voxelpose}$^\dag$ & 3 & $89.26$  & 93.91  & 24.02mm \\ 
Ours & 3 & \textbf{91.60} &  \textbf{94.14}   & \textbf{22.69mm} \\ 
\hline
    \end{tabular}}
    \label{tab:compare_panoptic}
\end{table}

\section{Network Architecture}

\begin{figure*}[th]
	\centering
	\includegraphics[width=0.99\textwidth]{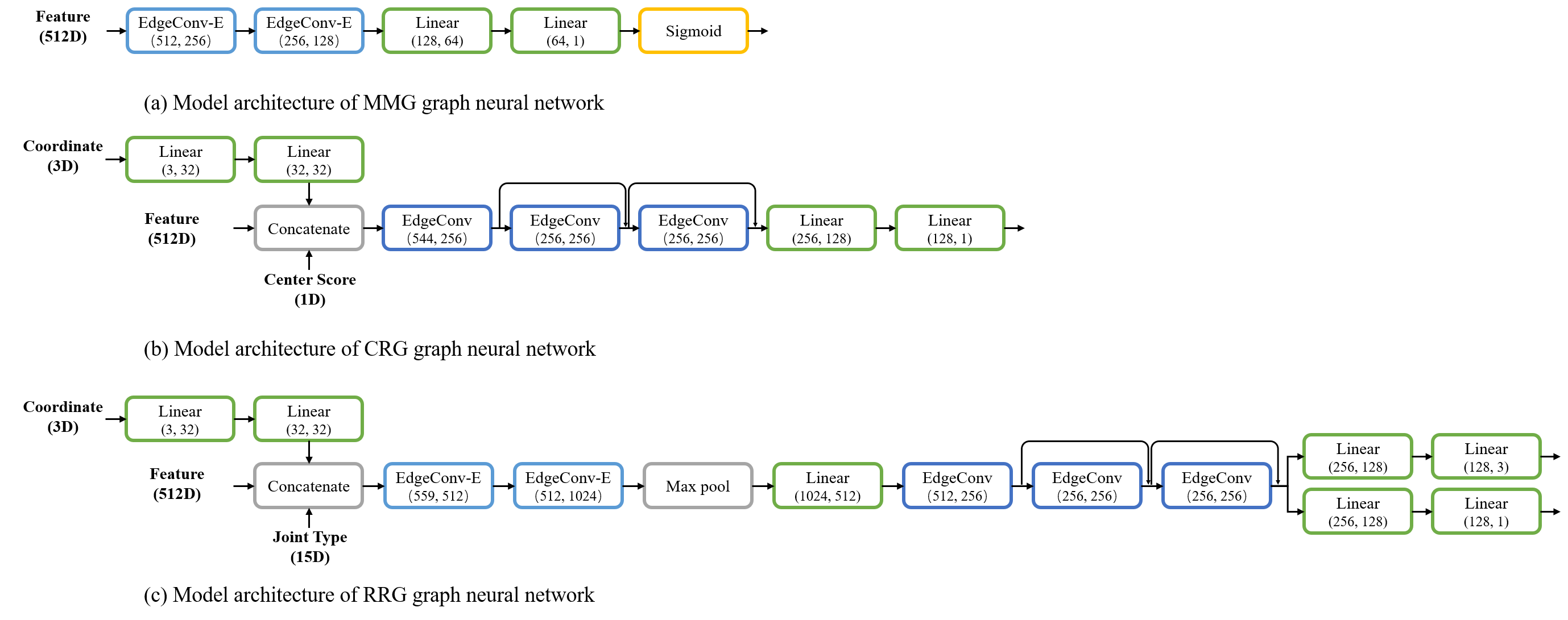}
	\caption{The model architectures of (a) MMG, (b) CRG and (c) PRG. `Linear' denotes the fully-connected layer, and `MaxPool' denotes the graph max-pooling layer. The input feature dimensions and the output feature dimensions are illustrated.}
	\label{fig:arch}
\end{figure*}

In this section, we illustrate the detailed graph model architectures of MMG, CRG and PRG in Figure~\ref{fig:arch}.

As shown in Figure~\ref{fig:arch} (a), the graph model of MMG consists of two layers of EdgeConv-E followed by two fully-connected (FC) layers. The input visual features $\mathbb{R}^{512}$ extracted in the image plane are first updated by the EdgeConv-E layers. Then the fully-connected layers, whose input is the concatenation of target vertex feature and relative source vertex feature, predict whether an edge is connecting 2D centers of the same person. 

As shown in Figure~\ref{fig:arch} (b), the input features come from three sources:
(1) the visual features $\mathbb{R}^{512}$ extracted in the image plane (2) the normalized 3D coordinates $\mathbb{R}^{3}$ of the query point (3) 2D center confidence score from the 2D backbone $\mathbb{R}^{1}$ . They are processed by fully-connected layers and then concatenated to produce a feature vector $\mathbb{R}^{545}$ for each vertex. The features are then processed by three layers of EdgeConv for cross-view feature message passing. A max-pooling layer is used for feature fusion and fully-connected layers to predict the center confidence score. To facilitate training, we adopt residual connections in between the EdgeConv layers.

As shown in Figure~\ref{fig:arch} (c), the input features also come from three sources: (1) the visual features $\mathbb{R}^{512}$ extracted in the image plane (2) the normalized 3D coordinates $\mathbb{R}^{3}$ of each joint in the initial pose (3) one-hot feature of the joint type $\mathbb{R}^{15}$. They are processed by fully-connected layers and concatenated to produce a feature vector $\mathbb{R}^{559}$ for each vertex. The features are then processed by two layers of EdgeConv-E for cross-view message passing. Then a max-pooling layer is applied to aggregate the cross-view features and coarsen the graph. The max pooled features are updated by the following three EdgeConv layers via effective information flow between the body joints. Similar to CRG, we add some residual connections to help model training. Finally, the extracted features are passed to two parallel MLPs (multi-layer perceptrons) to respectively regress a refinement vector and predict a confidence score for each joint. Both MLPs are composed of two fully-connected layers.

\section{Qualitative Comparisons}

In this section, we present more qualitative comparisons with Tu \etal~\cite{tu2020voxelpose} on the CMU Panoptic dataset (a, b, c) and the Shelf dataset (d).

\begin{figure*}[ht]
	\centering
	\includegraphics[width=0.88\textwidth]{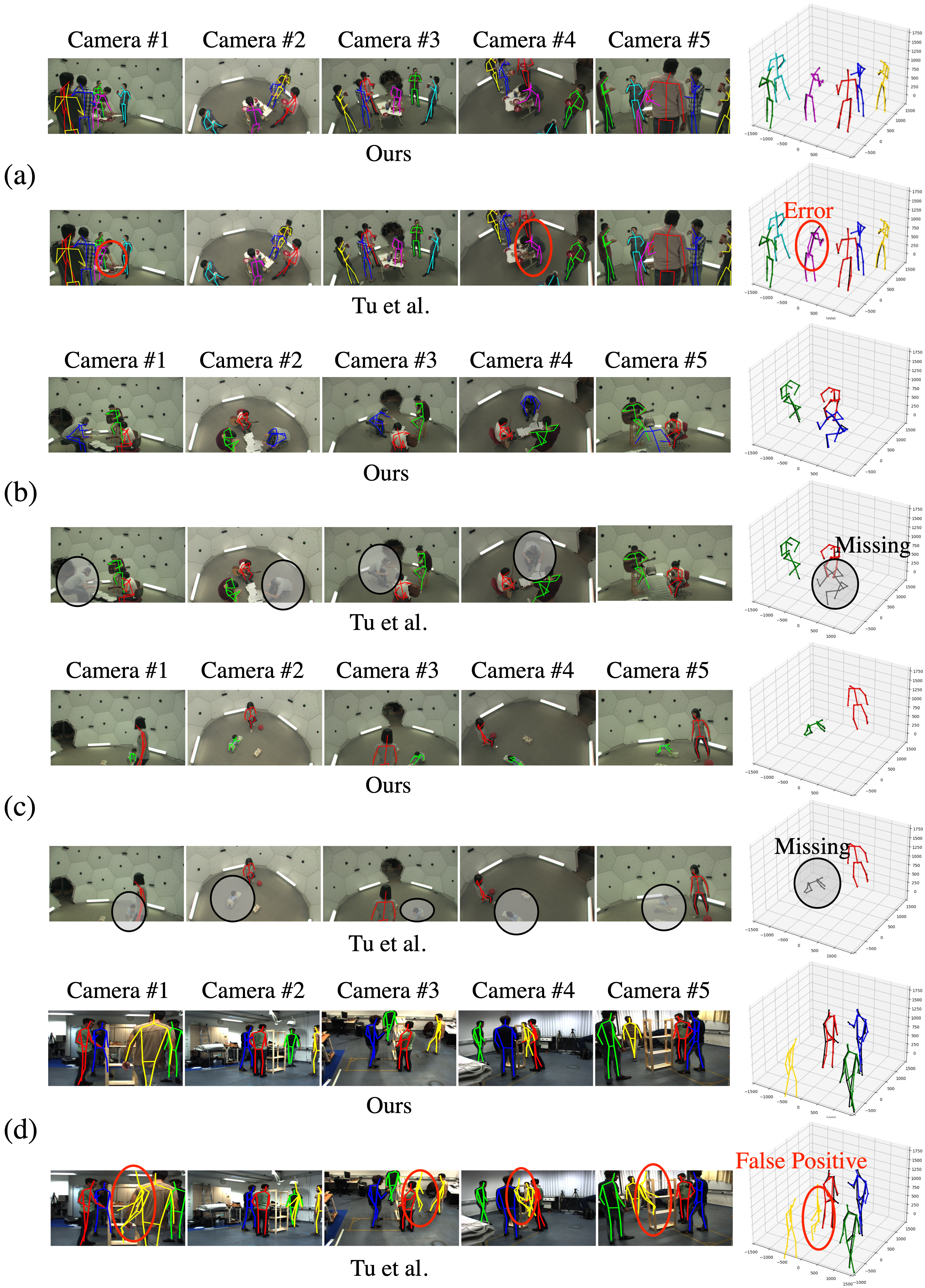}
	\caption{\textbf{Qualitative analysis on CMU Panoptic dataset (a, b, c) and Shelf dataset (d).} Estimated 3D poses and their 2D projections of ours, and Tu \etal~\cite{tu2020voxelpose}. The ground-truth 3D poses are in black, while the predicted 3D poses are in other colors (red, blue \etc). Inaccurate poses, false negatives, and false positives are highlighted with circles. Best viewed in color.}
	\label{fig:supp_cmu}
\end{figure*}

As shown in Figure~\ref{fig:supp_cmu} (a), the arm of the man (purple) is only visible in two camera views, and is occluded by other people or by himself in most views. This results in large 3D pose errors for Tu \etal~\cite{tu2020voxelpose}. Our proposed PRG can fix such kinds of pose errors, by considering both the geometric constraints and the human body structural relations. 

As shown in Figure~\ref{fig:supp_cmu} (b), many joints of the man (blue) are self-occluded by his own body in many camera views. This makes the visual features unreliable, leading to false negatives (FN) for Tu \etal~\cite{tu2020voxelpose}. In comparison, our proposed MMG and CRG learn to detect human centers in a coarse-to-fine manner via GCN. We are able to obtain more robust human detection results. 
As shown in Figure~\ref{fig:supp_cmu} (c), accurately predicting the poses of the little child (green) is challenging, due to insufficient training data. This example indicates that our proposed approach has better generalization ability towards rare poses.

As shown in Figure~\ref{fig:supp_cmu} (d), there is a false positive pose in the red circle estimated by Tu \etal~\cite{tu2020voxelpose}. In comparison, our approach achieves better performance and gets fewer false positives. Our proposed CRG together with PRG can suppress these false positives, by considering the multi-view features as a whole via GCN. 

\end{document}